\documentclass[a4paper,twoside]{article}

\usepackage{epsfig}
\usepackage{subfigure}
\usepackage{calc}
\usepackage{amssymb}
\usepackage{amstext}
\usepackage{amsmath}
\usepackage{amsthm}
\usepackage{multicol}
\usepackage{pslatex}
\usepackage{apalike}
\usepackage{hyperref}
\usepackage{gensymb}
\usepackage[table]{xcolor}	
\usepackage{SCITEPRESS}     

\begin{document}

\title{Building Robust Industrial Applicable Object Detection Models Using Transfer Learning and Single Pass Deep Learning Architectures}

\author{\authorname{Steven Puttemans, Timothy Callemein and Toon Goedem\'{e}}
\affiliation{KU Leuven, EAVISE Research Group, Jan Pieter De Nayerlaan 5, Sint-Katelijne-Waver, Belgium.}
\email{\{steven.puttemans, timothy.callemein, toon.goedeme\}@kuleuven.be}
}

\keywords{Deep Learning, Object Detection, Industrial Specific Solutions}

\abstract{The uprising trend of deep learning in computer vision and artificial intelligence can simply not be ignored. On the most diverse tasks, from recognition and detection to segmentation, deep learning is able to obtain state-of-the-art results, reaching top notch performance. In this paper we explore how deep convolutional neural networks dedicated to the task of object detection can improve our industrial-oriented object detection pipelines, using state-of-the-art open source deep learning frameworks, like Darknet. By using a deep learning architecture that integrates region proposals, classification and probability estimation in a single run, we aim at obtaining real-time performance. We focus on reducing the needed amount of training data drastically by exploring transfer learning, while still maintaining a high average precision. Furthermore we apply these algorithms to two industrially relevant applications, one being the detection of promotion boards in eye tracking data and the other detecting and recognizing packages of warehouse products for augmented advertisements.}

\onecolumn \maketitle \normalsize \vfill

\section{\uppercase{Introduction}}
\label{sec:introduction}

\noindent Several drawbacks have kept deep learning in the background for quite the while. Until recently deep learning had a very high computational cost, due to the thousands of convolutions that had to process the input data from a pixel level to a more content-based level. On high-end systems, reporting processing speeds of several seconds on VGA resolution has long been the state-of-the-art. This limited the use of these powerful deep learning architectures in industrial situations such as real-time applications and on platforms with limited resources. Furthermore deep learning needed a dedicated and expensive general purpose graphical processing unit (GPGPU) and an enormous set of manually annotated training data, both things that are almost never available in industrially relevant applications. The available frameworks for deep learning lacked proper documentation and guidelines, while pre-built deep learning models were not easily adaptable to new and unseen classes.

These issues no longer exist, and deep learning made a major shift towards usability and real-time performance. With the rise of affordable hardware, large public datasets \textit{(e.g. ImageNet \cite{deng2009imagenet})}, pre-built models \textit{(e.g. Caffe model zoo \cite{jia2014caffe})} and techniques like transfer learning, deep learning took a step closer towards actual industrial applications. Together with the explosion of stable and well documented open-source deep learning frameworks \textit{(e.g. Caffe \cite{jia2014caffe}, Tensorflow \cite{abadi2016tensorflow}, Darknet \cite{darknet13})}, deep learning opened up a world of possibilities.

\begin{figure}[b]
	\centering
    \includegraphics[width=0.48\textwidth]{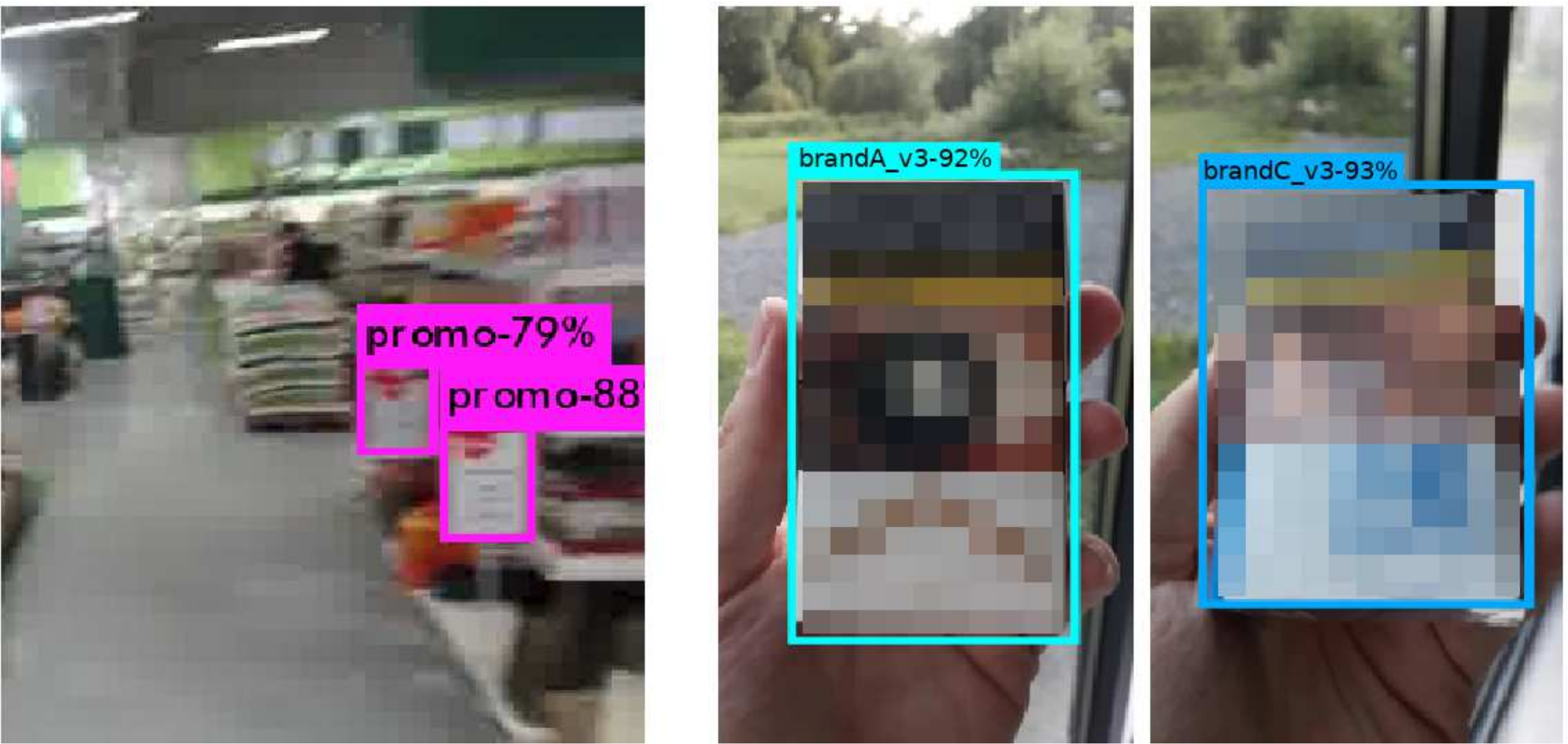}
    \caption{Examples of both applications: \textit{(left)} promotion boards and \textit{(right)} product packs detectors.}
    \label{fig:cases}
\end{figure}

Many industrially relevant applications do not focus on existing classes from academically provided datasets. Gathering tremendous amounts of training data and providing manual labelling, is a time-consuming and thus expensive process. Luckily transfer learning \cite{bengio2012deeptransferlearning} provides possibilities in this case. Given a limited set of manually annotated training samples, it retrains existing off-the-shelf deep learning models and adapts the existing weights so that the model can detect a complete new object class.

We apply transfer learning in two applications, while focussing on using a small set of training data for the new classes \textit{(to prove we do not need huge datasets for this task)} and simultaneously ensuring real-time performance. The first industrial relevant application handles the detection of promotion boards in eye-tracking data, allowing companies to analyse shopping behaviour, as seen in Figure \ref{fig:cases}(left). The second application handles the detection and recognition of warehouse products for augmented advertisement purposes \textit{(as seen in Figure \ref{fig:cases}(right))}. Specifically we look at a generic product box detector, followed by a multi-class 14-brand-specific detector.

The remainder of this paper is organized as follows. In section \ref{sec:relatedwork} we discuss related work and highlight the state-of-the-art in deep object detection. Section \ref{sec:dataset} discusses the collected dataset for training and validating the object detection models. This is followed by section \ref{sec:approach} discussing the selected deep learning framework and model architecture, used for transfer learning. Section \ref{sec:cases} handles each application case in detail. In section \ref{sec:results} we discuss the obtained average precisions and execution speeds. Finally we conclude this paper in section \ref{sec:conclusion} and provide some future work in section \ref{sec:futurework}.

\vspace{-2mm}
\section{\uppercase{Related Work}}
\label{sec:relatedwork}

\noindent Since convolutional neural networks process image patches, doing a multi-scale and sliding-window based analysis takes a lot of time, especially if image dimensions increase. Combined with the fact that deeper networks achieve a better detection accuracy, deep networks like \textit{DenseNet201} and \textit{InceptionV3} can easily take a couple of seconds for VGA resolution. To increase execution speed, academia introduced the concept of region proposal techniques. These are fast lightweight filters that pre-process images, looking for regions that might contain objects we are looking for. Instead of supplying the classification pipeline with millions of candidate windows, the region proposal algorithms reduce this to several thousands of windows, clearly reducing processing time.

Region proposing convolutional neural networks (R-CNN) \cite{ren2015faster} use a separate shallow region proposal network on the GPU that uses mutual information \textit{(some convolutional layers are shared)} of the actual detection pipeline to avoid unnecessary overhead. This enabled to run deep deep models like VGG19 at 10FPS for VGA resolution, giving a 30x speed improvement towards the original implementation. The `Single Shot Multibox'-detector (SSD) \cite{liu2016ssd} introduces an algorithm for detecting objects in image using only a single deep neural network, immediately grouping the convolutional output activations and returning object boxes. The `You Only Look Once'-detector (YOLO) \cite{redmon2016you} implements a similar approach, using anchor points that allow learned aspect ratios around pixel areas that have a high response. Both detectors remove the need of extra region proposal networks and perform bounding box prediction and class probability generation in a single run through the network.

If your industrial application concerns a class that is previously trained on any given public dataset \textit{(e.g. pedestrians, cars, bicyclists)}, these algorithms provide off-the-shelf pre-trained detection models. However, when your object class does not occur in any of the pre-trained networks, one needs a huge dataset and a lot of processing time to come up with a new deep learned model. That is why \cite{yosinski2014transferable} investigated the transferability of deep learning features and discovered that the first convolutional layers of any deep learning architecture actually stick to very general feature representations, and that the actual power resides in the linking of those features by finding the correct convolutional weights for each operator. This allowed applying the concept of transfer learning \cite{dauphin2012unsupervised} onto deep learning. By initializing the weights of the convolutional layers that provide a general feature description, using the weights of any pre-trained model, and by using a small number of annotated application-specific training samples to update the weights of all layers for the new object class, one can learn a complete new model. Combined with data augmentation \cite{gan2015learning}, a small set of these training samples can introduce enough knowledge for fine-tuning an existing deep model onto this new class.

\vspace{-2mm}
\section{\uppercase{Dataset and Framework}}
\label{sec:dataset}

\noindent For this paper we created two datasets used for training and evaluating our deep learned object detection models. In both cases we provided manual annotations, allowing for a precision-recall based evaluation.

The first dataset exists of several videos of an eye-tracker experiment in a Belgian warehouse. Data is collected by costumers, approached at the entrance and asked to go shopping wearing an eye-tracker, without explaining them the goal of the experiment. Two classes of promotion boards \textit{(small\_sign and large\_sign)} are manually annotated in each frame of the eye-tracker videos. As training data we use respectively 75 and 65 frames containing the promotion boards. As validation data, we use 420 and 960 frames respectively, containing both frames with and without the actual promotion boards.

\begin{table}[t]
  \centering
  \caption{Number of product box samples per brand.}
  \label{tbl:cig_dataset}
  \begin{tabular}{|c|c|c|}
  \hline
  \cellcolor{gray!25} \textbf{Label}         & \cellcolor{gray!25} \textbf{Training} & \cellcolor{gray!25} \textbf{Validation} \\
  \hline
  brandA\_v1          & 63                       &  163 \\
  \hline
  brandA\_v2         & 14                       &  149 \\
  \hline
  brandA\_v3         & 29                       &  157 \\
  \hline
  brandA\_v4        & 14                       &  167 \\
  \hline
  brandB\_v1      & 75                       &  157 \\
  \hline
  brandB\_v2     & 15                       &  222 \\
  \hline
  brandC\_v1            & 15                       &  185 \\
  \hline
  brandC\_v2             & 15                       &  188 \\
  \hline
  brandC\_v3            & 15                       &  229 \\
  \hline
  brandD\_v1                & 46                       &  183 \\
  \hline
  brandD\_v2               & 30                       &  146\\
  \hline
  brandD\_v3               & 15                       &  179 \\
  \hline
  brandE\_v1 & 15                       &  178 \\
  \hline
  brandE\_v2    & 15  						&  179 \\
  \hline
  \cellcolor{gray!25} \textbf{TOTAL:}       & \cellcolor{gray!25} \textbf{376}                &  \cellcolor{gray!25} \textbf{4279} \\
  \hline
  \end{tabular}
\end{table}

The second industrial relevant dataset contains 376 images of product packages. In each image the location of the package is manually annotated using a rectangular bounding box. In total 14 brands are included, allowing us to both provide labels of a product box and the associated brand, useful for training multi-class object detection models. A separate validation set is collected, existing of twentysix fifteen-second videos \textit{(@30FPS)}, containing different views of similar product boxes. Table \ref{tbl:cig_dataset} contains a detailed overview of the specific amounts of training and validation samples per brand.

As framework we decide to use Darknet \cite{darknet13}, a lightweight deep learning library, based on C and CUDA, which reports state-of-the-art performance for object detection using the YOLOv2 architecture. Moreover, it reaches inference speeds up to 120 FPS and integrates well with our existing C++ based software platform \textit{(using OpenCV3.2 \cite{bradski2000opencv})}. The framework does not require explicit negative training samples but rather uses areas in the image that are not labelled as object. We build our customized version of the framework, which is publicly available \footnote{\url{https://gitlab.com/EAVISE/darknet}} and allows integrated evaluation of trained models using the precision-recall metric.

\vspace{-2mm}
\section{\uppercase{Suggested Approach}}
\label{sec:approach}

\noindent In order to converge to an optimal configuration, deep learned models need large amounts of annotated training data and a lot of processing time. For the suggested YOLOv2 architecture, this is also the case. The default network configuration uses 800.000 floating point operations per given image. To optimize the weights assigned to each operation, one literally needs multiple thousands of training samples. For industrial applications this is not manageable, certainly if for each new object detection model, you need new and annotated training data. Transfer learning bridges the gap here. For our detection models we start from a YOLOv2 architecture previously trained on the Pascal VOC 2007 dataset with pre-trained weights. To be able to apply transfer learning onto a new object class several adaptations have to be made to the network.

\begin{enumerate}
\item We physically change the architecture of the network at the level of the class predictions and class probabilities. To ensure the convolutional layers output the correct format for the detection layer, we change the total number of filters in the final convolutional layer to $(N_{classes} + 5) \times N_{anchors}$. This allows the detection layer to convert the final layer activations into useful detections.
\item We adapt the amount of anchor ratios to our class specific problem. This ensures that in the detection layer, prediction boxes are generated fitted in a ratio that agrees to our actual fine-tuning data from our new object class.
\item We initialize the weights of all convolutional layers except the last one and leave the weights of the final convolutional layer and the detection layer uninitialized. This allows us to learn class specific weights for these deciding layers.
\end{enumerate}

Keeping these changes in mind, we transfer learn a new object model for our new object class. We allow all weights to update \textit{(including the pre-assigned weights)} and thus to converge towards the optimal solution given our new training data. We also use data augmentation to ensure we enrich the small application specific annotated training set.

In our experience the detection model generally fine-tunes over multiple thousands of iterations, providing a batch of 64 samples at each iteration to the model for updating the weights and reducing the loss rate on those given samples. If our model trains for 10.000 iterations, the model needs more than 640.000 training samples. Given we only have around 400 annotated training samples available this would mean that each sample is shown over a thousand times to the network, risking to overfit the model to the training data. Data augmentation applies specific transformations to the image to generate a large set of data from a small set of training samples, effectively reducing the number of times a single sample is shown to the network and thus avoiding overfitting. 

\newpage
During our training we allow the following augmentations on our limited set of manually annotated training data:

\begin{itemize}
\item{Randomly rescaling \textit{(max.$\pm$30\%)} of the input size of the first convolutional layer, making the detector more robust for multi-scale detections.}
\item{For each training sample, the algorithm randomly decides to flip around the vertical axis.}
\item{Input images are converted to HSV and a maximally $\pm$10\% deviation on the hue value is applied.}
\item{The average saturation and exposure of the input image can deviate 50\% from the input value.}
\item{We allow the annotations of the training samples to jitter for 20\% in relation to the original size, but the cropped or moved annotation still has to contain 80\% of the original annotation area.}
\end{itemize}

In general we notice that any object model we train is able to converge towards a stable model overnight, maximally taking a full 24 hours, on a Titan X (Pascal) and with a default architecture input resolution of $416 \times 416$ pixels. We halt the training when the loss rate on the provided training samples seems to drop under 0.5, what for our applications always produces a robust detection model. Continuing the training from that point, lets say for example up to 50.000 iterations, only introduces training data overfitting and thus a model that loses its generalization properties on an unseen validation set.

\vspace{-3mm}
\section{\uppercase{Practical Cases}}
\label{sec:cases}

\noindent Before we move on to qualitative and quantitative evaluation of our object models in section \ref{sec:results}, we first elaborate on how we built object detection models for our specific industrially relevant applications.

\subsection{Promotion boards}
\label{sec:caseAdvertisement}

\begin{figure}[t]
	\centering
    \includegraphics[width=0.48\textwidth]{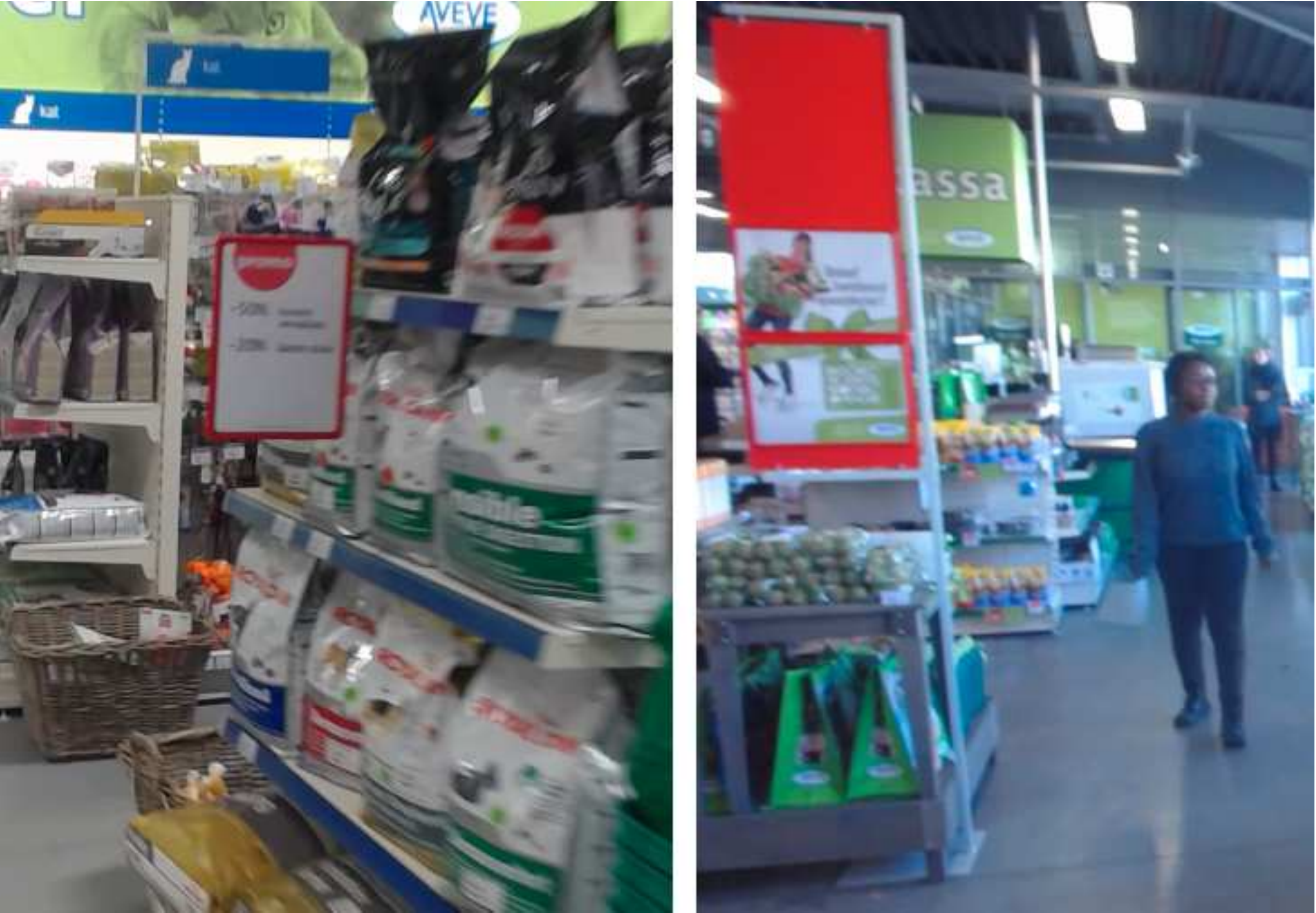}
    \caption{Two promotion board classes (left) small\_sign (right) large\_sign.}
    \label{fig:boards}
    \vspace{-2mm}
\end{figure}

For our promotion boards, we look into deep learning with two specific reasons:

\begin{enumerate}
\item We want to analyse complete eye-tracker experiments by linking object detections to eye-tracker data, removing the need for manual analysis.
\item We want to reduce the amount of annotations as much as possible without losing too much average precision on the obtained object detectors.
\end{enumerate}

We build two separate object detection models for this application. First we build a single-class detection model, able to detect the small red promotion board as seen in Figure \ref{fig:boards}(left). However, since the same shop also contains a second board of interest, seen as the large red promotion board in Figure \ref{fig:boards}(right), we explore the possibilities of building a two-class detector, able to detect and classify both signs at the same time and in a single run.

\subsection{Warehouse product packages}
\label{sec:caseCigarette}

\noindent Our second application is the robust detection and classification of warehouse product packages for augmented advertisements. In this application it is first of all our task to robustly locate the packages in any given input image. Furthermore we want to classify each product box in the same, single run of our deep learning classifier, and return its brand.

The packages all have general product box properties, while the print on the box distinguishes the brand. Therefore we start building a robust product package detector, able of localizing packages with a very high average precision in newly provided input images. This can easily be followed by a small classification network deciding which brand we have. However, doing this in a single run per image is our ultimate goal Once we succeed in building a robust product pack detector, we try pushing our luck , so we generate a 14-class brand detector, which is able to both localize and classify the product packs in a single run, completely removing the need of a separate classification pipeline.

\vspace{-3mm}
\section{\uppercase{Experiments and results}}
\label{sec:results}

\noindent This section discusses in more detail the achieved results with each trained object detector. Furthermore it highlights some of the issues we have when training these object detection models and discusses some of the limits of using the YOLOv2 architecture. Since all our models are fine-tuned from the same YOLOv2 architecture, we have an equal storage size of 270MB, while the GPU memory footprint of the same model equals 450MB.

\begin{figure}[t]
	\centering
	\includegraphics[width=0.49\textwidth]{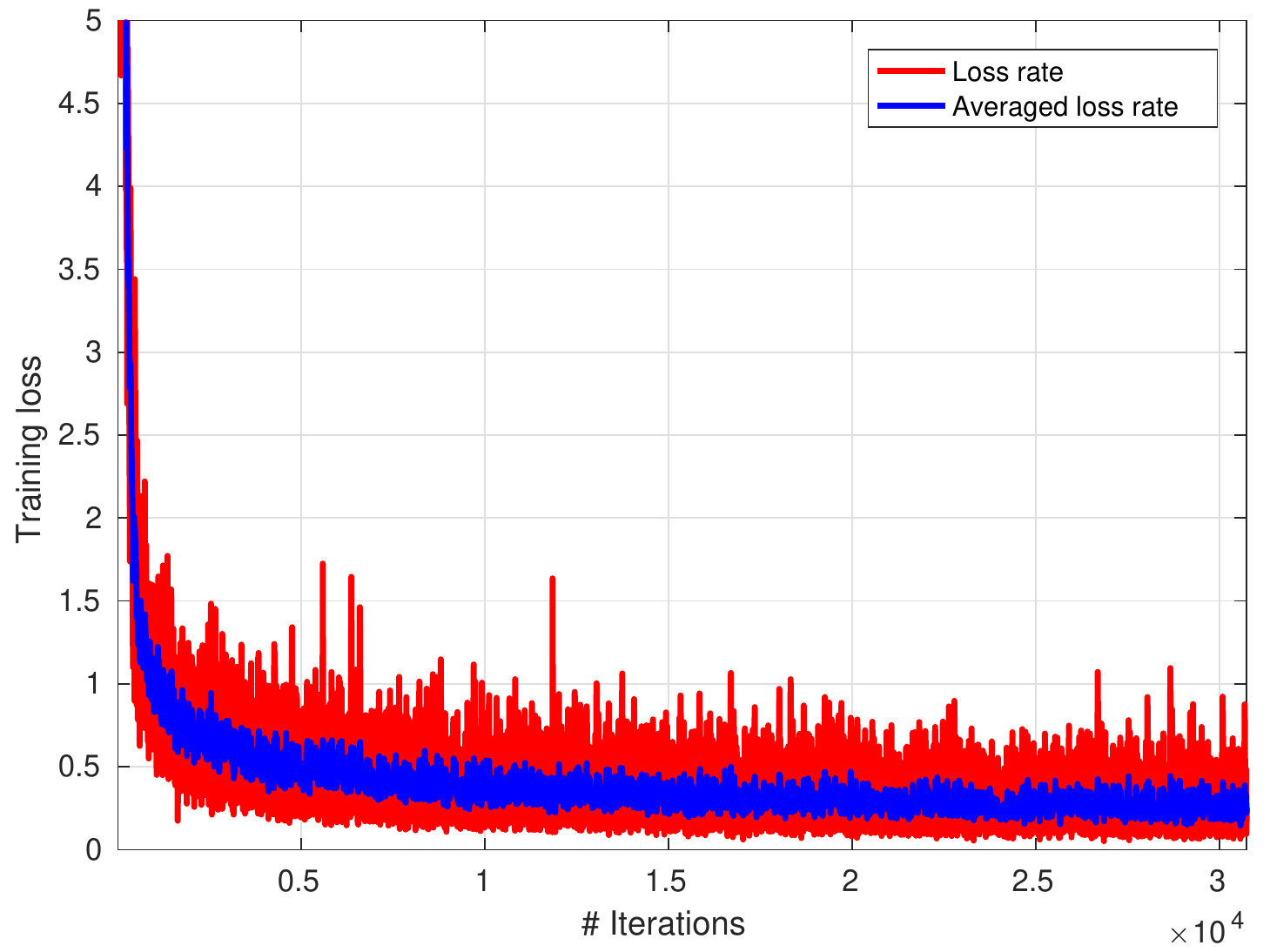}
    \caption{An example loss rate and average loss rate curve for single class advertisement board detector, in function of the iterations \textit{(with 64 sample batch per iteration)}.}
    \label{fig:loss_rate}
    \vspace{-2mm}
\end{figure}

\begin{figure}[b]
	\vspace{-3mm}
	\centering
	\includegraphics[width=0.49\textwidth]{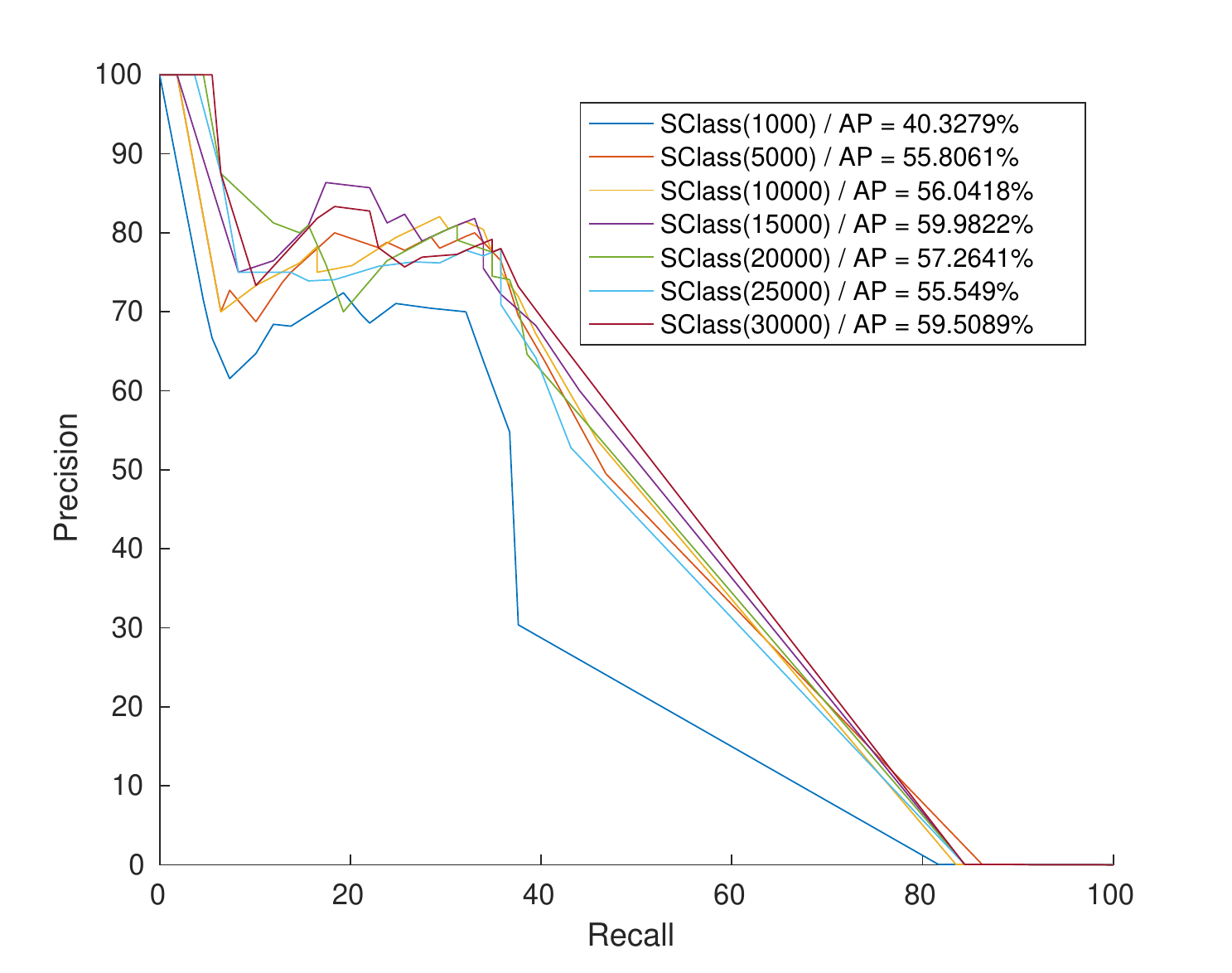} \hfill
    \includegraphics[width=0.49\textwidth]{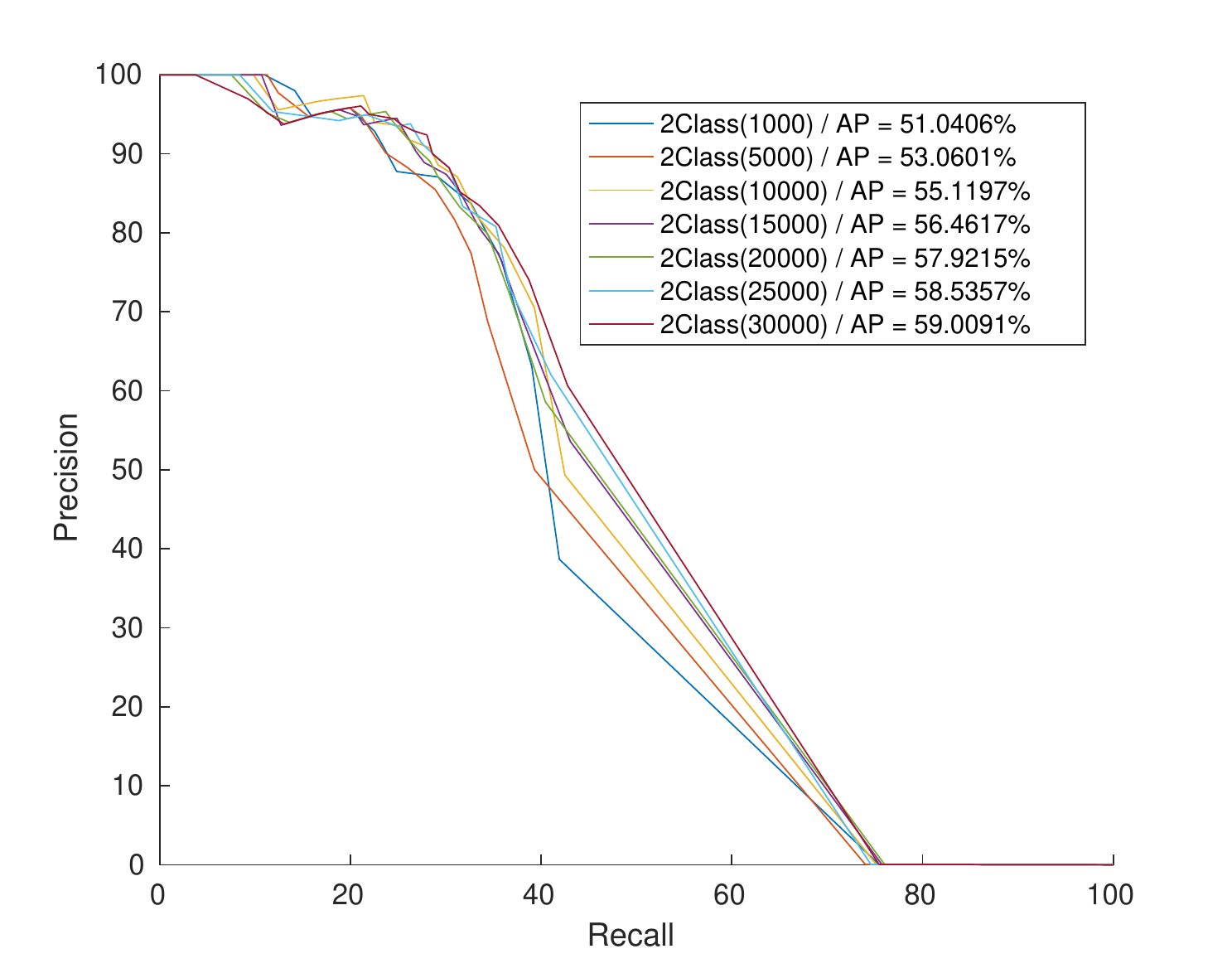} \hfill
    \caption{Precision-recall curves for the single class \textit{(top)} and the two-class \textit{(bottom)} object detection model.}
    \label{fig:pr_advertisement}
\end{figure}

\subsection{Stop the model transfer-learning}

\noindent We need a way of defining when our training algorithm reaches an optimal solution. At the same time we need to avoid overfitting the model to the training data, keeping a model that generalizes well to never seen before data. In order to check the training progress we continuously monitor the loss-rate on the given training data. This is monitored for every new batch of samples provided to the training algorithm, together with the average loss-rate over all batches. Once the average loss-rate no longer drops, we halt the training process.

An example of the loss rate tracking for one of our transfer-learned models can be seen in Figure \ref{fig:loss_rate}. In general we notice that a loss rate just below 0.5 seems optimal for any model, which is transfer-learned from the given YOLOv2 pre-trained architecture.

\subsection{Evaluate the promotion detectors}

Figure \ref{fig:pr_advertisement}(top) and \ref{fig:pr_advertisement}(bottom) shows the precision-recall curves for the trained promotion board models. For both models we run the training overnight halting the training at 30.000 iterations. We evaluate the precision, recall and average precision performance of each detection model at 1.000, 5.000, 10.000, 15.000, 20.000, 25.000 and 30.000 training iterations.

Initially we see an increase in average precision when raising the number of iterations for each model. However, once the average precision \textit{(calculated on the validation set)} starts dropping, we select the previous model as best fit, in order to keep the model that generalizes best on the given validation set. For the single-class detector this means a model of 15.000 iterations at an average precision of 59.98\% while for the two-class detector the 30.000 iterations model at an average precision of 59\% performs best. Looking at these curves it seems that the two-class model might still be converging to an optimal solution and thus continuing training with more iterations might be a feasible option.

We reckon our models seem to be under-performing, seeing their optimal configuration strands at about 65\% precision for a 50\% recall. This can be due to motion blur in the eye-tracker data in the validation set. The scales of the annotated validation data also differ from the training data, leading to unlearned object dimensions. Finally YOLOv2 has problems with detecting objects that are relatively small in comparison to the image dimensions, while this is not an issue for a human annotator.

However these precision and recall values are actually more than enough, given the context of the application. We only need to signal when a costumer has noticed a promotion board, related to the gaze-cursor location. If we are not able to find a promotion board on a smaller scale, once the costumer walks towards the sign, we are able to robustly detect it with a high probability. Looking at the execution speeds of the trained models, we notice that both the single- and two-class models perform around 55 FPS for a $1280 \times 720$ pixel resolution on a NVIDIA Titan X GPU. A video of the single class promotion board detector can be found at \url{https://youtu.be/dQIdRSDm6Jc}.

\subsection{Evaluating the product package and brand detectors}

For our general single-class product box detector, we perform transfer-learning for 5.000 iterations, while our fourteen-class brand detector is trained for 15.000 iterations. Both trainings lead to a convergence in loss-rate, just below the 0.5 threshold, previously indicated as a good stopping point.

\begin{figure*}[t]
	\centering
	\includegraphics[width=0.99\textwidth]{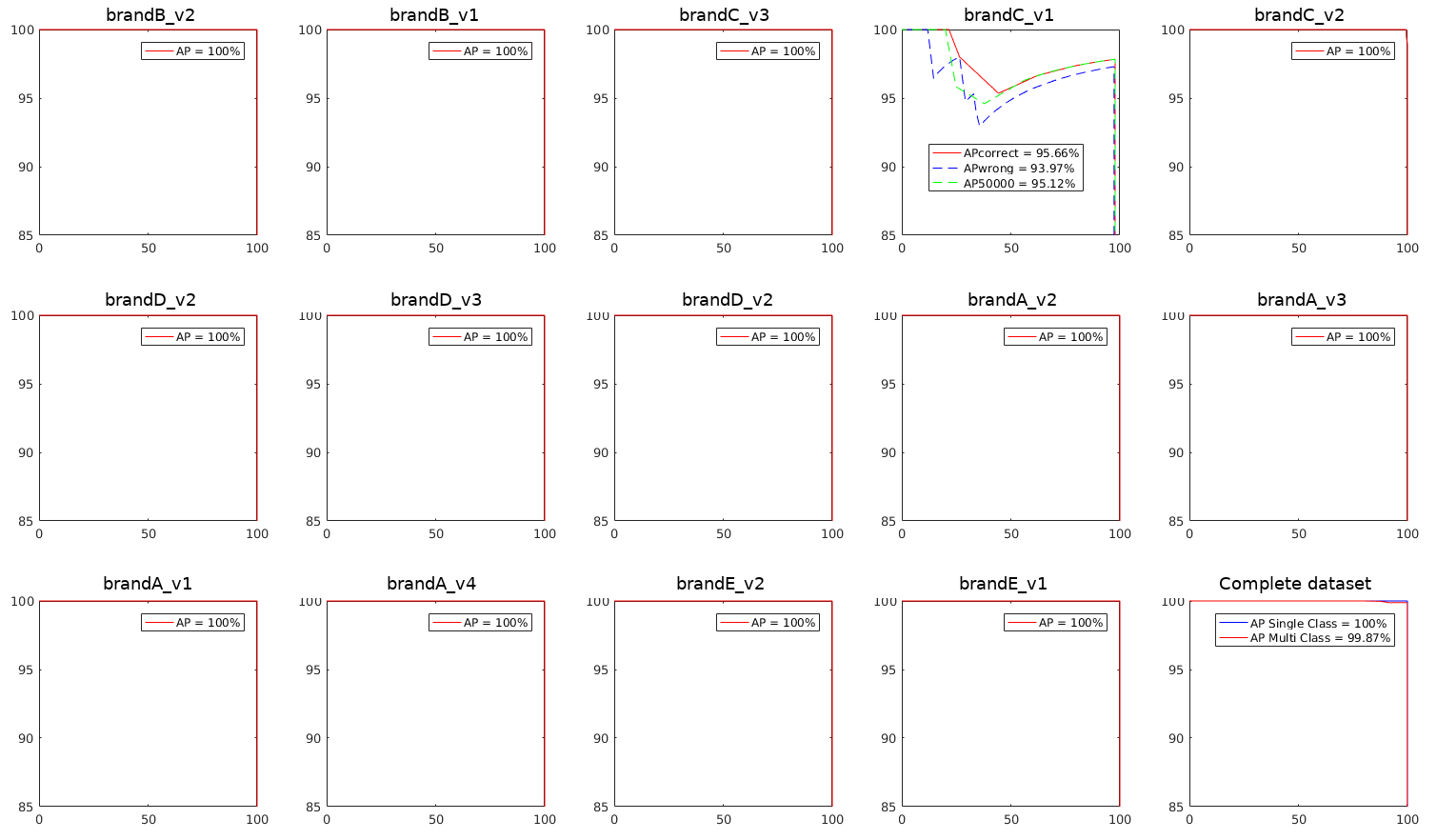}
    \caption{Precision-recall curves for our product brand case, showing a per (sub-)brand evaluation, a combined evaluation and a small extra investigation for the brandC\_v1 sub-brand.}
    \label{fig:overview_cigarettes}
\end{figure*}

Figure \ref{fig:overview_cigarettes} gives us an overview of the generated precision-recall curves for all the transfer-learned models. At the bottom left part of the figure we observe the performance on the complete validation dataset \textit{(so all fourteen brands together)}. Our general single-class product box detector is able to robustly detect every single object instance in the large validation set, obtaining an average precision of 100\%. The 14-class object detection model is as promising, obtaining an overall average precision of 99.87\%, directly providing the correct brand label in combination with an accurate localisation. 

The small difference in average precision between both models is caused by brandC\_v1, which is unable to reach a class-specific average precision of 100\%. We discovered four samples of the brandC\_v1 with a wrong label, which under data augmentation seems to have a significant influence. Before adjusting these labels, we obtained a 93.97\% average precision for this class, while afterwards we obtain a 95.66\% average precision.

\begin{figure}[t]
	\centering
	\includegraphics[width=0.49\textwidth]{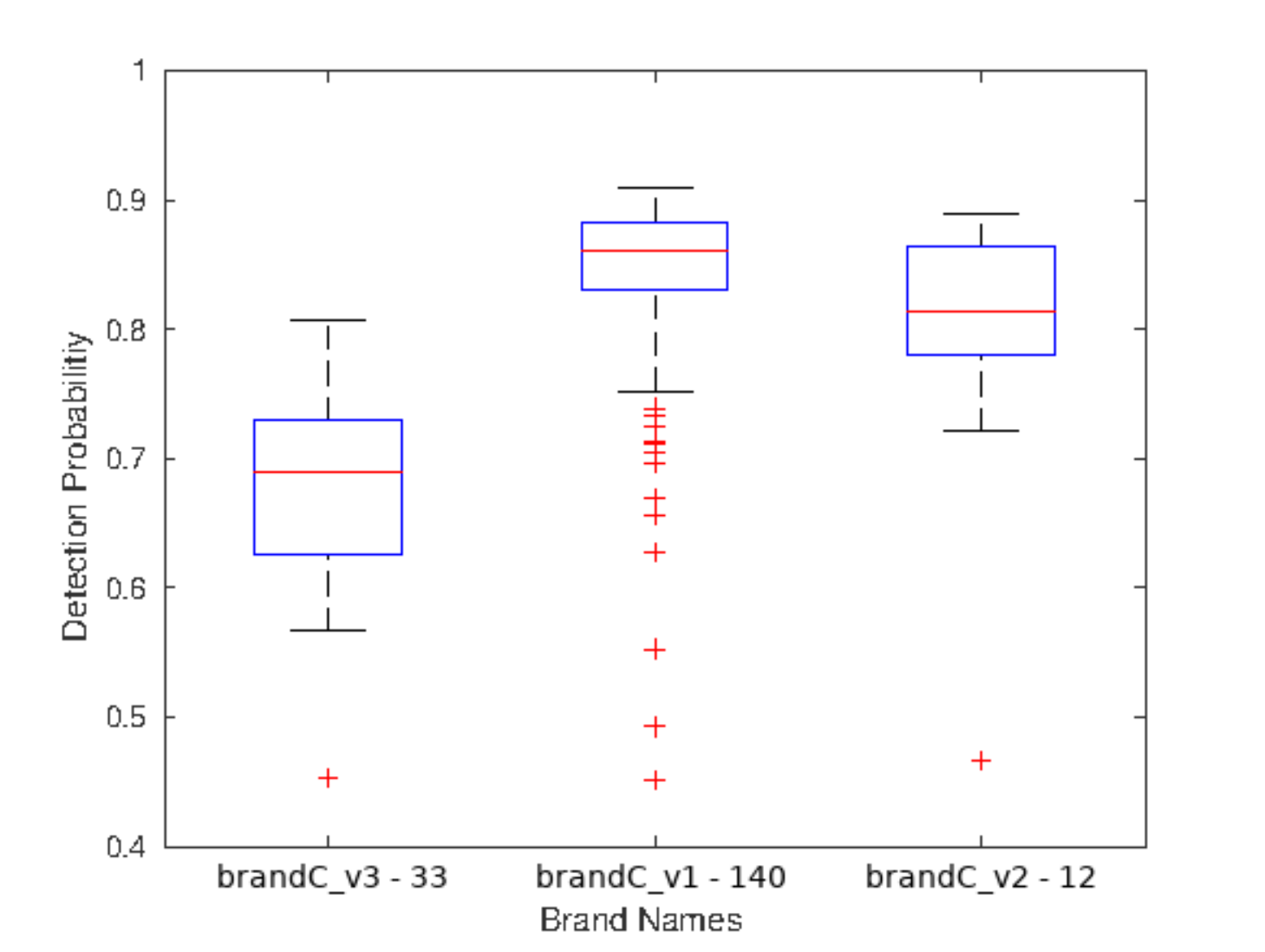}
    \caption{Box-plot of the brandC sub-brands probability scores indicating a significant number of v1 packages triggering v2 and v3 detections with a high probability.}
    \label{fig:boxplot}
\end{figure}

We compared the performance three different brandC classes \textit{(v1, v2 and v3)}, as seen in Figure \ref{fig:boxplot}, evaluated on brandC\_v1 validation data with the goal of uncovering where it goes wrong. There is still a large amount of validation data that gets classified as brandC\_v2 or brandC\_v3, and this with a high class probability, instead of providing the brandC\_v1 label. After visually inspecting the brandC class this behaviour is kind of expected, and actually the detectors are performing better than expected, since there are only very subtle visual changes between these three sub-brands. Keeping in mind that our network resizes the input data to a fixed $416 \times 416$ resolution this can lead to the loss of these very small details, resulting in a deep learning architecture that is even more challenged in finding an optimal separation between the sub-brand classes.

Looking at the execution speeds of the trained models, we notice that both the single-class and the fourteen-class product package detector are running at 70 FPS for a $720 \times 1280$ pixel resolution on an NVIDIA Titan X GPU.

\vspace{-2mm}
\section{\uppercase{Discussion and conclusions}}
\label{sec:conclusion}

\noindent In this section we want to discuss some restrictions of the current YOLOv2 framework, shed a light on some issues that still arise when using our models and finally draw some conclusions on our suggested approach. Finally we suggest ways for further improving the current pipeline.

\subsection{Drawbacks using YOLOv2}

Due to chosen internal construction of convolutional layers and the final detection layer \textit{(in which YOLO smartly combines it probability maps)}, it is unable to robustly detect objects that are small compared to the image dimensions. This happens quite frequently in our promotion board detection case, resulting in lower average precisions. Once the object instances cover more than 20-30\% of the image \textit{(e.g. the product box case)}, this is no longer an issue.

The grid based region proposal system in YOLOv2 forces neighbouring detections together in a single detection box, dropping the average precision. The architecture has a non-maxima suppression parameter that can be tweaked for this, but initial experiments show this does not resolve all issues.

Finally deep learning separates training data based on a combination of convolution filters, trying to find an optimal solutions. However we do not tell it how it should do its task or which features to ignore, which sometimes has undesired effects. E.g. in the case of our generic product box detector, when presented with a similar rectangular shape \textit{(e.g. a box of cereals or sugar)} the model also triggers a detection. This behaviour is expected since a product package is first of all a rectangular box.

\subsection{Avoiding detections on general rectangular shapes}

To remove the effect of other rectangular packages triggering the product package detector, we investigate the detection probability range. Given a set of product packs and some random packs with other aspect ratios, we noticed that there is a large probability gap between known and unknown packages. Simply placing a probability threshold of 65\% already results in the removal of all detections of non-known product packages and brands.

However we are interested in other possible solutions to this problem, that do not include setting a hard threshold on the probability output, since unseen data could screw up this approach and trigger detection probabilities for desired objects that are lower than the given threshold.

\subsubsection{Adding a non-standard box class to the multi-class detector}

\begin{figure}[t]
	\centering
	\includegraphics[width=0.49\textwidth]{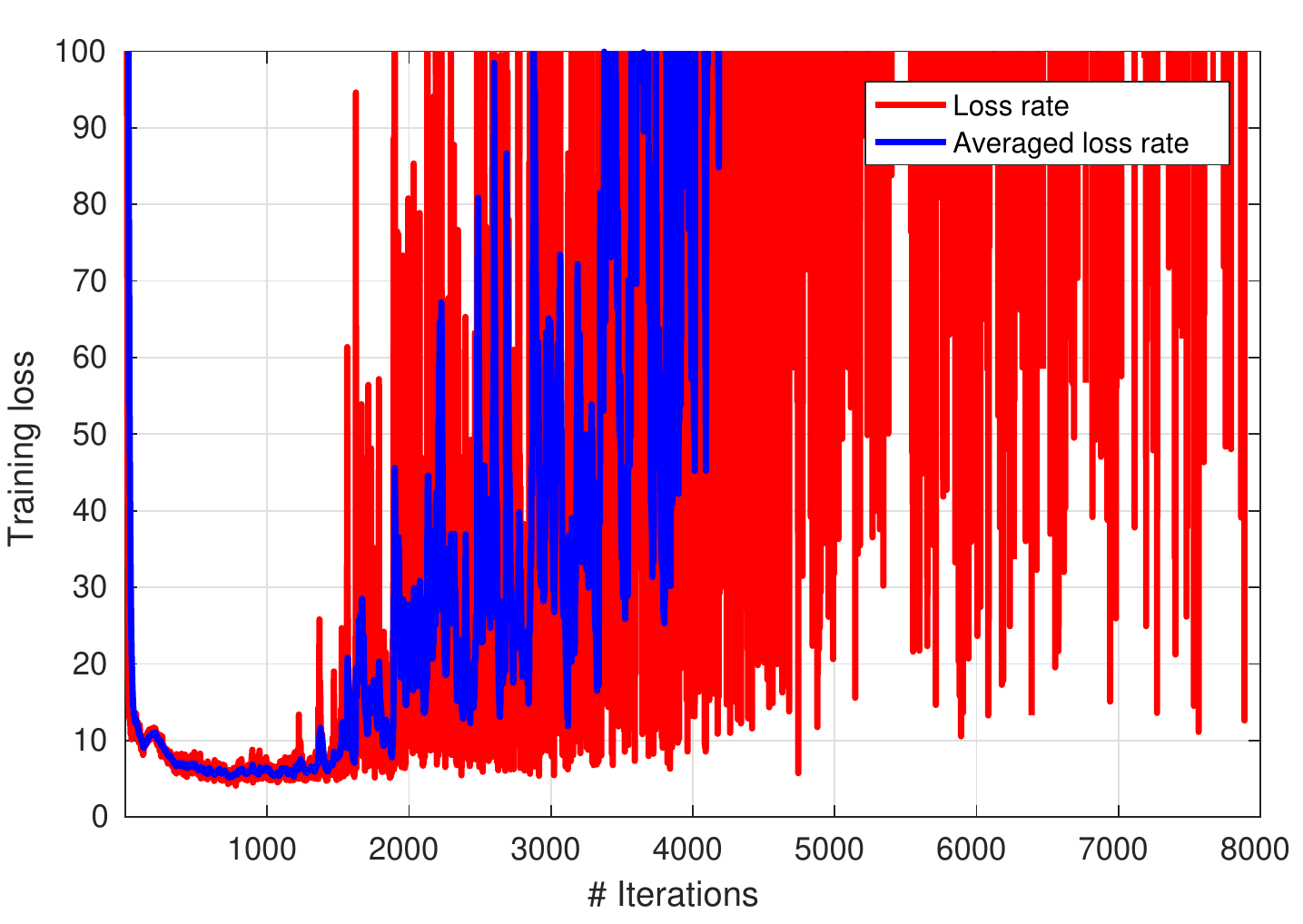}
    \caption{Training a brand-specific box detector with a generic box as extra class, leading to a loss rate explosion.}
    \label{fig:negative_class}
\end{figure}

By adding a class that contains all non-standard boxes we basically want to force the deep learning network to learn all product brands, and on top of that an extra generic box detector. We add extra training data, collected from 12 household rectangular boxes and started training again. Figure \ref{fig:negative_class} shows that the initial iterations show a promising drop in loss rate, however, at a certain moment we literally experience a training loss rate explosion, and the model is not able to recover from this. To be able to recover from this, a possibility might be to lower the learning rate from the moment the loss-rate increases again, since the giant loss-rates clearly show the architecture is on the wrong track looking for the optimal solution, given the multi-dimensional error surface.

\subsubsection{Adding hard negative training images}

A second attempt is made by adding hard negative training images, images containing generic unannotated box shapes, without changing the number of classes to learn. The architecture will force the learning process to try and make the response of those hard negative images zero and thus hopefully improve the ignoring power of the model for generic packages.

\begin{figure}[t]
	\centering
	\includegraphics[width=0.49\textwidth]{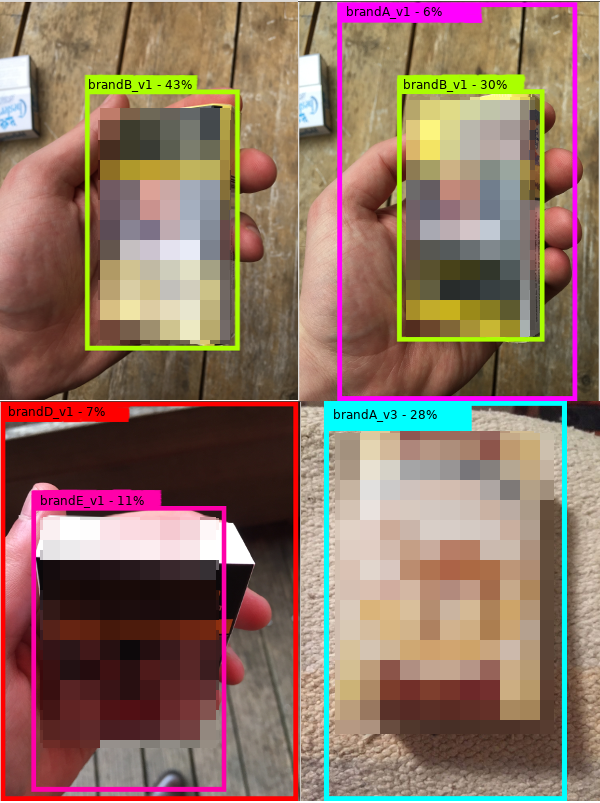}
    \caption{Example of classes that are unseen and untrained, yielding clearly lower probabilities allowing for easier separation based on class probability.}
    \label{fig:other_packages}
\end{figure}

This approach works and yields a transfer-learned model with a low loss-rate. The hard negative data is not removing the generic box detection properties from the model, which it still needs for detecting the actual brands, but when looking at the probabilities for the generic box detections, we clearly notice a larger gap in probability between object class instances and generic box detections.

Caution should be used here. This only works when using a limited set of hard negatives samples in relation to the amount of training samples per class. Adding too much hard negative samples increases the risk of generating  training sample batches that contain almost no actual object annotations, which forces to learn the model to detect nothing at all.

Examples of generic packages or packages of untrained brands, yielding low probabilities compared to actual trained brands can be seen in Figure \ref{fig:other_packages}. Knowing that actual trained classes trigger detections almost always above a probability of 85\% ensures us that there is still a margin for defining the most optimal threshold.

\subsection{Conclusion}

We proved that using deep learning for industrial object detection applications is no longer infeasible due to computational demands and unavailable hardware. By using an off-the-shelf deep learning framework and by using transfer learning we succeed in building several robust object detection models on an affordable general purpose GPU.

Our application of promotion board detection proves that the deep learned object detection pipeline is valid for industrially relevant cases, using only a very limited training set and thus minimal manual effort in labelling, yielding models that achieve moderate average precisions. At the same time we conclude that not all applications need a model with a 100\% average precision, in order to do its job, like in the case of automated eye-tracker data analysis.

Finally the application of product package detection and classification tries to push the limits of deep learning and model fine-tuning on industrial cases. By using again very limited datasets \textit{(keep in mind some classes have only 15 labelled image samples)} we achieve a perfect solution for our application, achieving a 100\% average precision when it comes to generic box detection. Furthermore if we directly incorporate the classification in the detection pipeline, we achieve a remarkable 99.87\% average precision. This perfectly proves that deep object detection could be a valid solution to several industrial challenges that are still around nowadays.

\subsection{Future Work}
\label{sec:futurework}

First of all we would like to do a deeper study of our promotion board detection case and how we can improve the moderate average precision rates obtained. We can only guess towards the actual reasons, but we are convinced that one of the issues could well be the sampling of the actual eye tracker video material, which contains huge amounts of motion blur, especially when people tend to move their head fast, for a quick look down the isle. Better eye-tracker hardware could be a possible solution here, or replacing the validation data with only clear eye-tracker data, removing the actual motion blurred evaluation frames. We also reckon our model will never be able to detect motion blurred advertisement boards, since we never used them as actual training data. Adding them to the actual training data might also improve the generalization capabilities of the deep learned model.

A challenge for the package detection and classification case exists in expanding the multi-class model. For now our model still needs an overnight training step. For most applications this is feasible, but there are still applications where this is not feasible at all. We should thus investigate how we can optimally expand existing models with an extra class, at a minimal processing cost. This research field is called incremental learning and already has several applicational fields, like object detection in video sequences, as described in \cite{kuznetsova2015expanding}.

Finally every augmented training sample is randomly flipped around the vertical axis in the Darknet data augmentation pipeline. While in most cases this helps to build robustness inside the detector, there are cases where this can worsens the actual model. We therefore suggest training models without these random flips, and comparing them to our already obtained results, to investigate their actual influence.

\vspace{-2mm}
\section*{\uppercase{Acknowledgements}}

This work is supported by the KU Leuven, Campus De Nayer and the Flanders Innovation \& Entrepreneurship (AIO).

\bibliographystyle{apalike}
{\small
\bibliography{VISAPP2017}}

\end{document}